\documentclass[letterpaper]{article} 
\usepackage{aaai25}
\usepackage{times}  
\usepackage{helvet}  
\usepackage{courier}  
\usepackage[hyphens]{url}  
\usepackage{graphicx} 
\usepackage{natbib}  
\usepackage{caption} 
\usepackage{algorithm}
\usepackage{algorithmic}

\usepackage{multirow}
\usepackage{graphicx}
\usepackage{booktabs}
\usepackage{amsfonts}
\usepackage{amsmath}
\usepackage{comment}

\usepackage{newfloat}
\usepackage{listings}
\DeclareCaptionStyle{ruled}{labelfont=normalfont,labelsep=colon,strut=off} 
\floatstyle{ruled}
\newfloat{listing}{tb}{lst}{}
\floatname{listing}{Listing}
\usepackage[hidelinks]{hyperref}
\nocopyright

\title{Integrating Audio, Visual, and Semantic Information for Enhanced Multimodal Speaker Diarization}
\author {
    Luyao Cheng\textsuperscript{\rm 1},
    Hui Wang\textsuperscript{\rm 1},
    Siqi Zheng\textsuperscript{\rm 1},
    Yafeng Chen\textsuperscript{\rm 1},
    Rongjie Huang\textsuperscript{\rm 2},\\
    Qinglin Zhang\textsuperscript{\rm 1},
    Qian Chen\textsuperscript{\rm 1},
    Xihao Li\textsuperscript{\rm 3},
}
\affiliations {
    \textsuperscript{\rm 1}Alibaba Group\\
    \textsuperscript{\rm 2}Zhejiang University\\
    \textsuperscript{\rm 3}University of North Carolina at Chapel Hill\\
    shuli.cly@alibaba-inc.com
}

\usepackage{bibentry}

\begin{document}

\maketitle

\begin{abstract}
Speaker diarization, the process of segmenting an audio stream or transcribed speech content into homogenous partitions based on speaker identity, plays a crucial role in the interpretation and analysis of human speech. Most existing speaker diarization systems rely exclusively on unimodal acoustic information, making the task particularly challenging due to the innate ambiguities of audio signals.
Recent studies have made tremendous efforts towards audio-visual or audio-semantic modeling to enhance performance.
However, even the incorporation of up to two modalities often falls short in addressing the complexities of spontaneous and unstructured conversations.
To exploit more meaningful dialogue patterns, we propose a novel multimodal approach that jointly utilizes audio, visual, and semantic cues to enhance speaker diarization. Our method elegantly formulates the multimodal modeling as a constrained optimization problem.
First, we build insights into the visual connections among active speakers and the semantic interactions within spoken content, thereby establishing abundant pairwise constraints. Then we introduce a joint pairwise constraint propagation algorithm to cluster speakers based on these visual and semantic constraints. This integration effectively leverages the complementary strengths of different modalities, refining the affinity estimation between individual speaker embeddings.
Extensive experiments conducted on multiple multimodal datasets demonstrate that our approach consistently outperforms state-of-the-art speaker diarization methods. 
\end{abstract}

%

\maketitle

\section{Introduction}

\begin{table*}[]
\caption{Advantages and disadvantages of different modalities for speaker diarization.}
\label{tab:modality}
\setlength\tabcolsep{12pt}
\begin{tabular}{ccc}
\toprule
Modality                 & Advantages & Disadvantages \\ \midrule
\multirow{2}{*}{Audio}   & Conveys direct voice characteristics.     &Vulnerable to environmental interference. \\
                         & Tracks speaker activity seamlessly.       &Fails in handling simultaneous speech. \\ \hline
\multirow{2}{*}{Vision}  & Offers distinctive visual cues.        & Cannot handle off-screen voices.             \\
                         & Robust to complicated acoustic conditions.    & Sensitive to camera quality, angle, and distance.             \\ \hline
\multirow{2}{*}{Text} & Identifies speaker-turn with semantic breaks.        & Sensitive to transcription errors.            \\
                         & Provides semantic context.        &  Often presents ambiguity regarding speaker identity.\\ \bottomrule
\end{tabular}

\end{table*}

In the fields of human-computer interaction (HCI) and human-robot interaction (HRI), addressing multi-party dialogue problem is frequently essential~\cite{DBLP:conf/emnlp/OuchiT16,DBLP:conf/acl/GuTLXGJ20}. In a conversation situation where multiple speakers are involved, one crucial challenge that must be tackled along with automatic speech recognition (ASR) and natural language processing (NLP) is to correctly recognize and assign temporal segments of speech or transcribed texts to corresponding speakers. 
In this procedure, human individuals rely on a combination of listening, observing, and understanding to easily pinpoint who is speaking at any given moment. This spontaneous process of deciphering speech underscores a growing desire for machines capable of executing the same task with high accuracy.
In the field of speech signal processing, this task is referred to as Speaker Diarization  aiming at determining "who spoke when" within an audio stream~\cite{Wang2018SD,8683892,Anguera2012Review,Anguera2007Beamforming,Tranter2006overview,DBLP:conf/icassp/ZhengHWSFY21}. 

Traditional unimodal speaker diarization approaches only rely on acoustic cues to differentiate between speakers~\cite{Sell2014SDivector,Park2020AutoTuningSC,landini2022bayesian,DBLP:conf/emnlp/DuZZY22}. These methods often suffer from low-quality acoustic environments, which are typically characterized by the presence of background noise, reverberation, and overlapping speech from multiple speakers~\cite{Reynolds2005audio,Park2022Review}.
In last decade, much research attempted to investigate joint modeling of multiple modality information from a variety of view points, enhancing the generalization and reliability of speaker diarization. 
~\cite{who-said-that,xu2022ava,chung2020spot,gebru2017spatiotemporalSD} exploit the audio-visual association, based on the assumption that a speech signal is typically synchronized with the facial attributes and lip motion of active speakers.
~\cite{flemotomos2022role,Park2018MultimodalSS,ZuluagaGmez2021BertrafficBJ} employed both lexical and acoustic features to identify roles in a specific two-speaker scenario, where the speakers assume distinct roles and are expected to follow different linguistic patterns.
~\cite{Cheng2023ExploringSI,DBLP:journals/corr/abs-2309-10456, Flem2020Linguisti} proposed to develop supplementary language sub-tasks to detect semantic change points within dialogues, aiming to provide guidance for the audio-only diarization process. 

While there are some known attempts to jointly model audio-visual or audio-textual information for speaker diarization, there is an obvious absence in addressing the comprehensive utilization of all three modalities, audio, visual, and textual information, in a unified framework. 
As summarized in Table~\ref{tab:modality}, each modality offers distinct and complementary strengths. Audio signals, being the primary source of speech content, provide direct access to vocal characteristics such as pitch, timbre, and speaking rate, which are essential for speaker identification. 
Visual cues can assist in distinguishing speakers by capturing unique facial features and tracking lip movements over time in low-quality acoustic environment. 
Textual data, transcribed from ASR modules, provide rich contextual and semantic content, which can reveal clear linguistic patterns and identify speaker-turns based on semantic breaks. 
Concurrently, the inherent limitations of each individual modality also constrain the efficacy of unimodal speaker diarization.
We believe that the complementary natures of all three modalities, when carefully combined,  can potentially yield a performance leap beyond the sum of their individual contributions.
To successfully integrate them under one unified framework, we introduce a clustering method based on constrained optimization. By carefully constructing visual and semantic constraints, we effectively incorporate multimodal information in the joint constraint propagation.


Specifically, we first utilize voice activity detection(VAD) to obtain active speech segments and extract speaker embeddings, similar to the prevalent audio-only speaker diarization systems. 
Then we introduce additional visual components such as face recognition and lip movement detection to obtain visual connections among speakers, establishing pairwise constraints on visually active speakers.  
Subsequently, text-based dialogue detection and speaker-turn detection models are used to construct an understanding of semantic structures of the conversations. 
Finally,  a joint pairwise constraint propagation algorithm is introduced to estimate a refined affinity matrix of speaker embeddings and facilitate speakers clustering process. 
In this framework, multimodal modeling is explicitly formulated as a constrained optimization problem.

To comprehensively evaluate the effectiveness of our method, we conducted experiments using multiple multimodal datasets. These included a video dataset that we collected and manually annotated, as well as several popular open-source datasets. The results indicate a significant performance improvement and a robust generalization capability of our method.


To the best of our knowledge, this paper is the first effort to systematically integrate three modalities - visual, audio, and semantic information to improve the performance of speaker diarization. This study represents contribution to the field of multimodal speaker diarization, enhancing the existing literature with richer modality information. Through this augmented approach, we aim to catalyze subsequent advancements and broaden the scope for future research in the domain.

The main contributions of this paper are as follows: 
\begin{itemize}
	\item We propose a novel framework to incorporate audio, visual and semantic information into speaker diarization, leveraging the complementary strengths of three modalities.
	\item We introduce a joint pairwise constraint propagation method into the speaker clustering process, effectively enhancing the understanding of conversational structure  from distinct modality perspectives.
    \item We contribute a video evaluation set that contains speaker identity labels, their corresponding speech activity timestamps, and speech content. This addresses the absence of comparable public benchmark for multimodal speaker diarization, providing a much-needed standard for performance assessment.
\end{itemize}



\begin{figure*}
  \centering
  \includegraphics[width=0.9\textwidth]{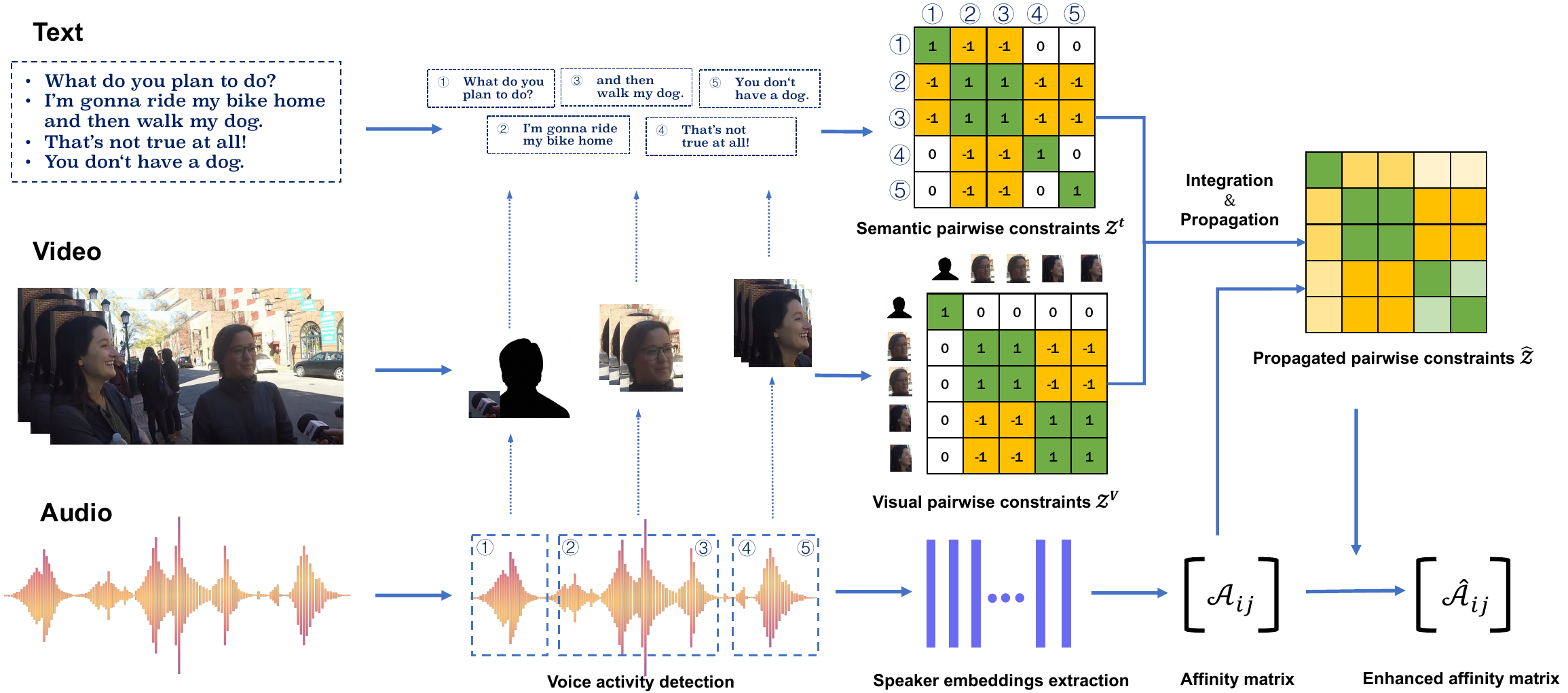}
  \caption{An overview of our proposed multimodal speaker diarization system. It incorporates additional visual and textual processing modules that independently extract visual and semantic constraints. By integrating and propagating knowledge derived from these different insights, comprehensive multimodal pairwise constraints are generated, serving as a robust guidance for enhancing the audio-based diarization.}
  \label{fig:overview}

\end{figure*}

\section{Related Work}

\subsection{Audio-only Speaker diarization}\label{sec:acoustic_only}
Audio-only speaker diarization has been studied extensively~\cite{Anguera2012Review,Park2022Review}. A typical speaker diarization systems employ a multi-stage framework~\cite{Ajmera2004Clustering,Sell2014SDivector,landini2022bayesian, Park2020AutoTuningSC, Zheng2022ReformulatingSD}, including voice activity detection (VAD)~\cite{Gelly2018VAD}, speech segmentation~\cite{Wei2022Turn}, acoustic embedding extraction~\cite{Sell2014SDivector,xvector, DBLP:conf/icassp/YuZSLL21,DBLP:conf/interspeech/ZhengLS20,DBLP:journals/corr/abs-2305-12838} and clustering~\cite{von2007tutorial,landini2022bayesian}. 
Recently, end-to-end neural diarization (EEND) where individual sub-modules in traditional systems can be replaced by one neural network has received more attention with promising results~\cite{Fujita2019Interspeech,Fujita2019ASRU,Horiguchi2020Attractors}. 
Due to lack of available large-scale data, EEND is usually trained on simulated dataset and suffers from generation issue in real-word applications. 
The transformer-based architecture of encoders and decoders causes computational inefficiencies when processing long audio sequences.
In general, it tends to be used combining with clustering-based diarization in most methods~\cite{Kinoshita2020IntegratingEN,Keisuke2021Advances}.

\subsection{Audio-visual Speaker diarization}
Facial activities and lip motion are highly related to speech~\cite{Yehia1998QuantitativeAO}. 
Visual information contains a strong clue for the identification of speakers and the location of speaker changes~\cite{Yoshioka2019Advances}, which can be used to significantly improve the accuracy of speaker diarization. Most methods leverage the audio and visual cues for diarization using synchronization between talking faces and voice tracks ~\cite{Ahmad2019MultimodalSD,Chung2018PerfectMI,xu2022ava}. 
Another work~\cite{who-said-that} adopted a synchronisation network to get self-enrolled speaker profiles and reformulate the task into a supervised classification problem. 
Recently, an interesting and promising direction is to use separate neural networks to process data streams of two modalities and directly output speech probabilities for all speakers simultaneously~\cite{He2022eendav}, similar to audio-only EEND frameworks.


\subsection{Audio-textual Speaker diarization}
Some previous works~\cite{ZuluagaGmez2021BertrafficBJ, flemotomos2022role, Park2018MultimodalSS, Paturi2023LexicalSE} utilized semantic information derived from transcription to estimate the role profiles and detect speaker change point, demonstrating improvement in specific role-playing conversations, such as job interviews and doctor-patient medical consultations.
Other works \cite{Kanda2021TranscribetoDiarizeNS, Wei2022Turn, Khare2022ASRAwareEN} enhanced ASR models to capture speaker identity through joint training of paired audio and textual data, which typically require substantial annotated multi-speaker speech data.
More recent works~\cite{Park2023EnhancingSD, Wang2024DiarizationLMSD, Cheng2023ExploringSI} employed large language models as post-processing to correct word speaker-related boundaries according to local semantic context.


\subsection{Pairwise Constrained Clustering}

As mentioned in Section~\ref{sec:acoustic_only}, speaker diarization systems often necessitate the introduction of an unsupervised speaker clustering algorithm, owing to the need to handle an indeterminate number of speakers. When introducing multimodal information, due to the inability to directly compare cross-modal information for similarity, it becomes essential to incorporate other modal information as weakly supervised signals into the speaker clustering; this process is known as constrained clustering~\cite{DBLP:journals/access/BibiAG23}.

Pairwise constrained clustering is one such classic methodology. Pairwise constraints (must-link and cannot-link) are well-studied and they provide the capability to define any ground truth set partitions~\cite{DBLP:conf/icml/DavidsonR07}. Since weakly supervised signals often do not encompass all target samples for clustering, various pairwise constraint propagation methods~\cite{Lu2011ExhaustiveAE} have been proposed to increase the number of pairwise constraints from a limited number of initial ones. Initially confined to data mining datasets~\cite{DBLP:conf/icml/HoiJL07}, the application of pairwise constrained clustering has expanded into multimodal areas such as vision and text~\cite{DBLP:conf/mm/YangHLCW14, DBLP:journals/pami/YanZYH06}. Advancing with theoretical progress, pairwise constraint propagation algorithms have increasingly integrated complex optimization techniques, including Non-negative Matrix Factorization (NMF)\cite{DBLP:journals/cvm/Fu15}, Inexact Augmented Lagrange Multiplier (IALM)\cite{DBLP:conf/mm/LiuJHZ19}, and deep learning outcomes~\cite{DBLP:journals/corr/abs-2101-02792, DBLP:journals/datamine/ZhangZBD21}.

\section{Methods}

The proposed framework's overview is depicted in Figure~\ref{fig:overview}. 
We will introduce joint pairwise constraint propagation in Sec.~\ref{jpcp}, visual constraints construction in Sec.~\ref{vision}, and semantic constraints construction in Sec.~\ref{text}, separately. 


\subsection{Joint Pairwise Constraint Propagation with multimodal Information}
\label{jpcp}
Considering that the audio contains comprehensive speaker-related information over time, we employ audio-based models, specifically a VAD model and a speaker embedding extractor, to obtain a sequence of acoustic speaker embeddings $E = \{e_1, e_2, ..., e_N | e_i \in \mathbb{R}^D \}$, by applying sliding windows to the audio data.
Subsequently, we compute the affinity matrix $\boldsymbol{\mathcal{A}} = \{\mathcal{A}_{ij}\}_{N \times N}$, where $\mathcal{A}_{ij} = g(e_i, e_j)$ and $g(\cdot)$ represents the similarity measurement.
It should be noted that this affinity matrix may contain errors resulting from acoustic environmental interference. 

Assuming we have access to speaker-related cues from additional sources of information, we can derive various types of constraint pairs: must-link $\mathcal{M}$ and cannot-link $\mathcal{C}$, defined as:
\begin{equation}\label{eq:def_link}
	\begin{aligned}
		\mathcal{M}^k &= \{(e_i, e_j) | l(e_i) = l(e_j)\}, \\
		\mathcal{C}^k &= \{(e_i, e_j) | l(e_i) \neq l(e_j)\},
	\end{aligned}
\end{equation}
where $l(\cdot)$ denotes the speaker label associated with an acoustic speaker embedding, and $k$ is the index of sources type.
For different modality information, the criteria for establishing $\mathcal{M}$ and $\mathcal{C}$ is different, which will be described in Sec.~\ref{vision} and Sec.~\ref{text} according to specific situation.
Then each constraint is initially encoded into a matrix $\boldsymbol{\mathcal{Z}}^{k}$:
\begin{equation}\label{eq:constraint_mat}
	\mathcal{Z}^{k}_{ij} = \begin{cases}
		+1 & \text{if } (e_i, e_j) \in \mathcal{M}^k, \\
		-1 & \text{if } (e_i, e_j) \in \mathcal{C}^k, \\
		0 & \text{otherwise}.
	\end{cases}
\end{equation}

A series of constraint matrix $\boldsymbol{\mathcal{Z}}^{k}$ are integrated into a final constraint matrix $\boldsymbol{\mathcal{Z}}$. During the integration process, some scenarios are relatively straightforward. For instance, if an embedding pair $(e_i, e_j)$ belongs to $\bigcap_k \mathcal{M}^k$, then $(e_i, e_j)$ is considered as a must-link constraint pair. Conversely, if $(e_i, e_j)$ resides in $\bigcap_k \mathcal{C}^k$, it is a cannot-link constraint pair due to agreement between all modalities. However, there are evidently more complex scenarios, where the constraint matrices conflict with one another, such as $(e_i, e_j) \in (\mathcal{M}^1 \cap \mathcal{C}^2)$ or $(e_i, e_j) \in (\mathcal{M}^2 \cap \mathcal{C}^1)$. To address these issues, we introduce acoustic information as the arbiter in the final determination. To summarize, we compute the integrated constraint scores following the given formula:
\begin{equation}
    \begin{aligned}
        \boldsymbol{\mathcal{Z}}' = \sum_k{\alpha_k \boldsymbol{\mathcal{Z}}^k} + \beta \boldsymbol{\mathcal{A}} - \theta
    \end{aligned}
\end{equation}
where $\alpha_k, \beta$ represent the weight hyper-parameters for different modalities, and $\theta$ is the bias. Then, $\boldsymbol{\mathcal{Z}}'$ is converted into a binarized constraint matrix $\boldsymbol{\mathcal{Z}}$ according to a threshold $\delta$.
\begin{equation}
    \begin{aligned}
        \mathcal{Z}_{ij} = \begin{cases}
		+1 & \text{if } \mathcal{Z}'_{ij} > \delta, \\
		-1 & \text{if } \mathcal{Z}'_{ij} < -\delta, \\
		0 & \text{else}.
        \end{cases}
    \end{aligned}
\end{equation}

The constraint matrix $\boldsymbol{\mathcal{Z}}$ may be sparse. Constraint information is confined to discrete and localized regions. It is essential to deploy a constraint propagation algorithm to efficiently broadcast the constraint information in $\boldsymbol{\mathcal{Z}}$ on a larger scale. Specifically, we employ E2CP~\cite{Lu2011ExhaustiveAE} algorithm to obtain propagated constraints  $\boldsymbol{\hat{\mathcal{Z}}}$:
\begin{equation}\label{eq:propagation}
	\boldsymbol{\hat{\mathcal{Z}}} = (1 - \lambda)^{2}(\mathbf{I} - \lambda\mathbf{L}_e)^{-1}\boldsymbol{\mathcal{Z}}(\mathbf{I} - \lambda\mathbf{L}_e)^{-1},
\end{equation}
where $\mathbf{L}_e = \mathbf{D}_e^{-1/2}\boldsymbol{\mathcal{A}}\mathbf{D}_e^{-1/2}$ is the normalized Laplacian matrix, and $\mathbf{D}_e$ is the degree matrix of $\boldsymbol{\mathcal{A}}$ and $\mathbf{I}$ is a identity matrix. The parameter $\lambda \in [0, 1]$ modulates the impact degree of the constraints. The refined affinity matrix $\boldsymbol{\hat{\mathcal{A}}} \in \mathbb{R}^{N \times N}$ is then updated to incorporate the influences of the propagated constraints $\boldsymbol{\hat{\mathcal{Z}}}$:
\begin{equation}\label{eq:change_affinity}
	\hat{\mathcal{A}}_{ij} = \begin{cases}
		1 - (1 - \hat{\mathcal{Z}}_{ij})(1 - \mathcal{A}_{ij}) & \text{if } \hat{\mathcal{Z}}_{ij} \ge 0, \\
		(1 + \hat{\mathcal{Z}}_{ij}) \mathcal{A}_{ij} & \text{if } \hat{\mathcal{Z}}_{ij} < 0.
	\end{cases}
\end{equation}

Upon calculating the affinity matrix $\boldsymbol{\hat{\mathcal{A}}}$, it is then fed into the clustering process to derive the ultimate speaker diarization results.
It is worth noting that there is no limit to the number of constraint types $k$. We can extract diverse constraint matrices related to different modal data. These constraint matrices can be perceived as prior knowledge, directing the clustering attention toward a particular view of the data. 
In this paper, we fix k at 2, thereby extracting two distinct constraint types: visual constraint $\boldsymbol{\mathcal{Z}^\mathit{v}}$ and semantic constraint $\boldsymbol{\mathcal{Z}^t}$.

\subsection{Visual constraints construction}
\label{vision}
The speaker-related visual constraints is constructed through the following steps, similar to~\cite{chung2020spot, xu2022ava}.

\textbf{Face tracking}. The initial phase involves detecting and tracking faces within the video content over time. A CNN-based face detector~\cite{Liu2018RFB} is used to continuously locate faces across frames, thereby creating a consistent track for each face present, using a position-based tracker. Only those face tracks that correspond with audible speech segments, as identified by VAD, are retained for further processing.

\textbf{Active speaker detection}. This step takes the cropped face video and corresponding audio as input and decides whether the tracked faces correspond to active speakers at any given moment. To achieve this, a two-stream network~\cite{tao2021someone}, consisting of temporal encoders and an attention-based decoder, is used to incorporate audio-visual synchrony and determine target speaker activity in the audio stream. In order to ensure the effectiveness of the subsequent process, we established a threshold to filter out video frames with relatively low confidence levels.

\textbf{Face clustering}. A face recognition CNN~\cite{Huang2020CurricularFace} is utilized to extract embeddings for every face track. The embeddings are extracted at uniform intervals, such as every 200 ms, in each face track. These are then clustered using Agglomerative Hierarchical Clustering (AHC).

By integrating these steps, constraints based on visual information are obtained. Faces clustered to the same speaker are considered as must-link constraints, while those clustered to different speakers are cannot-link constraints. 
Each face is aligned with respective acoustic embeddings $e_i$ along the time axis. 
If an acoustic embedding corresponds to multiple faces, we will select the speaker associated with the majority of those faces.
As a result, the obtained visual constraint $\boldsymbol{\mathcal{Z}^\mathit{v}}$ is consistent with the shape of the acoustic affinity $\boldsymbol{\mathcal{A}}$.

\subsection{Semantic constraints construction}
\label{text}

\begin{figure}
    \centering
    \includegraphics[width=0.47\textwidth]{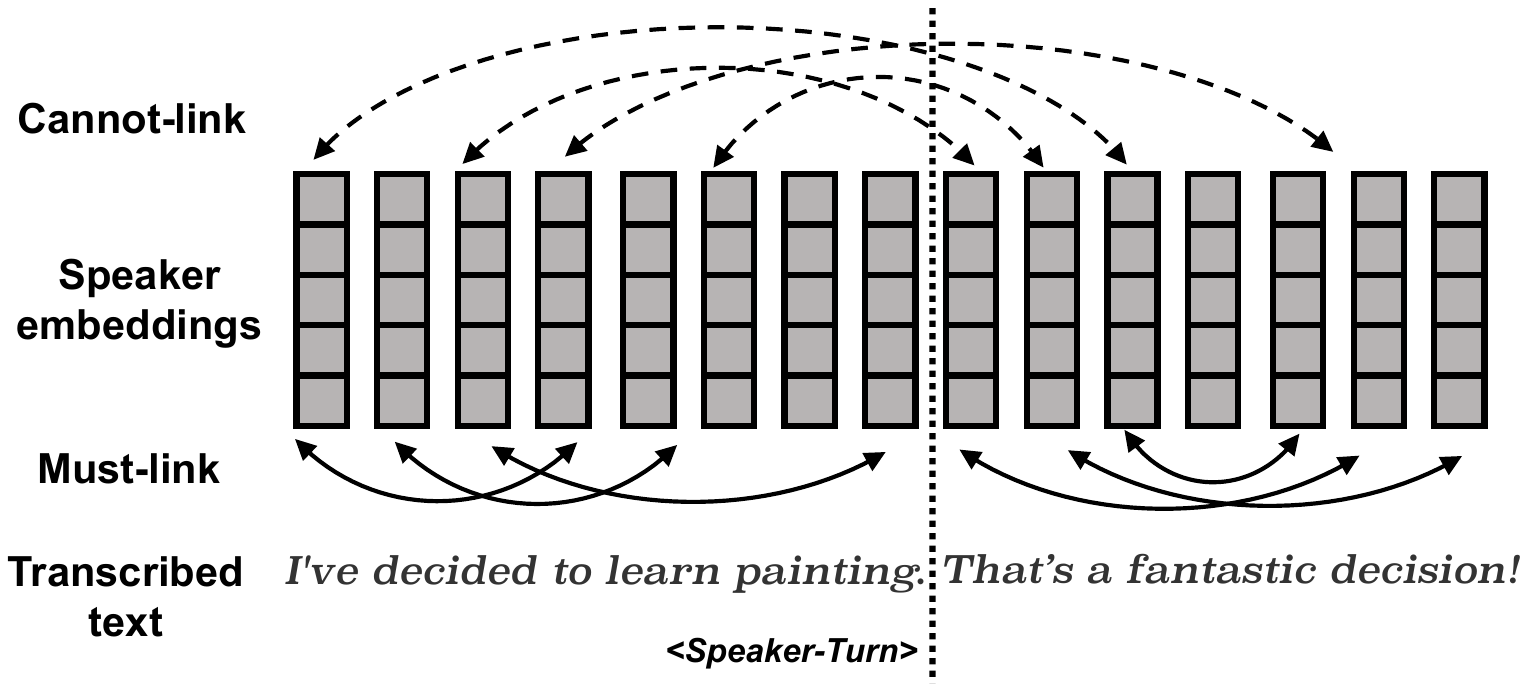}
     \caption{Semantic constraint construction based on dialogue detection and speaker-turn detection. Text segments judged as non-dialogue indicate that the associated embeddings are related through must-link constraints, depicted by solid connections below. Conversely, a detected transition point dictates that embeddings spanning this point should be connected with cannot-link constraints, as represented by dashed connections above.}
    \label{fig:pcc_constraints}
\end{figure}

To extract speaker-related information from the transcriptions, we constructed two Spoken Language Processing (SLP) tasks: (1) \textbf{Dialogue Detection} discriminates between multi-speaker dialogues and monologues, conceptualized as a binary classification challenge. (2) \textbf{Speaker-Turn Detection} assesses each sentence in a sequence to estimate speaker change, functioning as a sequence labeling problem that identifies semantically significant speaker role transitions.
Semantic constraints can be formulated based on the outputs of these two tasks.
Specifically, must-link $\mathcal{M}^t$ is formed between two embeddings if they are sourced from the same non-dialogue segment. Conversely, cannot-link $\mathcal{C}^t$ is established between embeddings separated by a detected speaker-turn boundary, as illustrated in Figure~\ref{fig:pcc_constraints}.

\begin{table*}
\centering
\caption{The results of speaker diarization with multimodal constraints}
\label{tab:constrained_sd}
\begin{tabular}{cccccccc}
\hline
\multirow{2}{*}{Cluster Algorithm} & \multirow{2}{*}{Constraints}  & \multicolumn{2}{c}{Cluster Metrics} & \multicolumn{4}{c}{Speaker Metrics(\%)}                                 \\ \cline{3-8} 
                                   &                               & ARI$\uparrow$              & NMI$\uparrow$              & DER$\downarrow$             & JER$\downarrow$              & TextDER$\downarrow$         & CpWER$\downarrow$            \\ \hline
VBx                                & No Constraints                & 0.903           & 0.893           & 10.31         & 29.28          & 4.23          & 18.03          \\
SC                                 & No Constraints                & 0.918           & 0.899           & 9.37          & 27.21          & 3.25          & 17.04          \\
E2CP + SC                          & Semantic Constraints          & 0.924           & 0.904           & 9.12          & 25.98          & 3.02          & 16.86          \\
E2CP + SC                          & Visual Constraints            & 0.924           & 0.905           & 9.13          & 26.02          & 3.02          & 16.83          \\
E2CP + SC                          & Semantic + Visual Constraints & \textbf{0.925}  & \textbf{0.908}  & \textbf{9.01} & \textbf{22.57} & \textbf{2.89} & \textbf{16.36} \\ \hline
\end{tabular}%
\end{table*}

\begin{table*}[]
\centering
\caption{Constraints derived from various modalities. We separately evaluated the accuracy and coverage of these constraints.}
\label{tab:constraint_constructions}

\begin{tabular}{ccccccc}
\hline
\multirow{2}{*}{Constraints} & \multicolumn{3}{c}{Accuracy(\%)} & \multicolumn{3}{c}{Coverage(\%)} \\ \cline{2-7} 
                                  & Must-Link  & Cannot-Link & Total & Must-Link  & Cannot-Link  & Total  \\ \hline
Semantic Constraints              & 99.75      & 84.80       & 99.40     & 1.23       & 0.08         & 0.49      \\ \hline
Visual Constraints                & 99.07      & 97.87       & 99.32     & 22.81      & 21.78        & 22.53      \\ \hline
Semantic + Visual Constraints     & 99.11      & 97.83       & 99.34    & 23.65      & 21.84        & 22.87      \\ \hline
\end{tabular}%

\end{table*}

There is an inherent transitivity associated with must-link constraints. That is, if $(e_i, e_j) \in \mathcal{M}^t$ and $(e_j, e_k) \in \mathcal{M}^t$, then it can be inferred that $(e_i, e_k) \in \mathcal{M}^t$. Unfortunately, such a property does not extend to cannot-link constraints. If $(e_i, e_j) \in \mathcal{C}^t$ and $(e_j, e_k) \in \mathcal{C}^t$, we cannot ascertain the relationship between $e_i$ and $e_k$, as in a real meeting scenario. Following a dialogue between speakers A and B, either speaker C may begin speaking, or speaker A may continue. Hence, semantic constraints derived solely from aforementioned methods can only determine the relationship between embeddings adjacent to a speaker-turn, and cannot influence embeddings separated by extended temporal intervals.



\section{Experiments}
\subsection{Datasets}

To evaluate our methods, we have constructed a diverse evaluation video dataset sourced from in-the-wild scenarios. This dataset includes dialogues featuring 2 to 10 English-speaking participants from interviews, talk shows, meetings, press conferences, round-table discussions, and TV shows, totaling approximately 6.3 hours of content. Individual videos range from 7 to 29 minutes and have been meticulously annotated with speaker identities, speech activity timestamps, and content. We intend to release this dataset publicly to advance multimodal research.

In addition to our evaluation dataset, we conducted supplementary experiments using three public multimodal datasets: AVA-AVD~\cite{xu2022ava}, AIShell-4~\cite{aishell42021}, and Alimeeting~\cite{alimeeting2022}. The AVA-AVD dataset, which focuses on the multimodal analysis of audio-visual diarization, includes over six languages and provides rich scenarios and face annotations; however, it lacks ground truth transcripts. In contrast, AIShell-4 and Alimeeting are both Mandarin datasets that offer speaker-labeled transcripts, making them particularly suited for various speech tasks. The combination of these varied datasets further validates the effectiveness and applicability of our methods across different domains.

\subsection{Implementation Details}
In our system, the audio-based diarization modules follow the pipeline outlined in \cite{Cheng2023ExploringSI}. 
Our speaker embedding extractor is an adaptation of CAM++ \cite{cam++}\footnote{The pretrained CAM++ came from \url{https://github.com/modelscope/3D-Speaker}}, which has been trained on VoxCeleb dataset~\cite{voxcelebNagraniCXZ20}.
For the visual componets, we employ a series of pretrained models for different tasks: 
RFB-Net~\cite{Liu2018RFB}\footnote{The pretrained RFB-Net came from \url{https://github.com/Linzaer/Ultra-Light-Fast-Generic-Face-Detector-1MB}} for face detection, 
TalkNet~\cite{tao2021someone}\footnote{The pretrained TalkNet came from \url{https://github.com/TaoRuijie/TalkNet-ASD}} for active speaker detection, 
and CurricularFace model~\cite{Huang2020CurricularFace}\footnote{The pretrained CurricularFace model came from \url{https://modelscope.cn/models/iic/cv_ir101_facerecognition_cfglint}} for extracting face embeddings.
To transcribe audio into text, we utilize the ASR model, Paraformer~\cite{Gao2022ParaformerFA}, which has been trained with the aid of the FunASR~\cite{Gao2023FunASRAF} toolkits\footnote{The ASR and forced-alignment models came from \url{https://github.com/modelscope/FunASR}}. 
These off-the-shelf pretrained models, crucial to our system, are all accessed through open-source platforms.


For semantic tasks, we pretrained muliple models with open-source meeting datasets for different scenarios. Specifically, we employed AMI~\cite{DBLP:conf/mlmi/CarlettaABFGHKKKKLLLMPRW05}, ICSI~\cite{Janin2003TheIM} and CHiME-6~\cite{Watanabe2020CHiME6CT} to generate English semantic models, and used Alimeeting and AIShell-4 training datasets to obtain Mandarin semantic models.
In our experiments, a sliding window strategy was employed, featuring a window size of 96 words and a shift of 16 words, to construct training sets for dialogue detection and speaker-turn detection training from transcripts with speaker annotations within these datasets. 
All that training was conducted using a pre-trained BERT model~\cite{Devlin2019BERTPO}. 
Subsequently, we employed the methods described in Section~\ref{text} to construct the semantic constraints.


The VBx approach~\cite{landini2022bayesian} represents a canonical method for speaker diarization, where the original paper employs speaker embeddings based on the x-vector model. We substituted this with the more robust CAM++ model. Furthermore, given that the post-processing step of the E2CP method integrates spectral clustering (SC)~\cite{von2007tutorial}, we also explored the performance of a method utilizing solely speaker embeddings and SC. The aforementioned two audio-only methods will serve as the baselines for this study.



As introduced in Section \ref{jpcp}, after obtaining multimodal pairwise constraints, our clustering process is divided into two submodules: constraint propagation and post-cluster. We employ E2CP as the core algorithm for constraint propagation. When only visual constraints are present, the parameter $\lambda$ in E2CP is set to 0.8, while it is set to 0.95 when semantic constraints are incorporated. For the post-cluster phase, we adhered to the SC algorithm consistent with the baseline. Our method, inspired by the work presented in the paper~\cite{Park2020AutoTuningSC}, incorporates refinement operations such as row-wise thresholding and symmetrization to enhance the performance of spectral clustering. In the row-wise thresholding of SC, the p-percentile parameter is set to 0.982. To account for the randomness of the clustering algorithm, our reported clustering metric is the average of 10 repetitions.


\subsection{Evaluation Metrics}
We report two popular clustering algorithm metrics: Normalized Mutual Information (NMI)~\cite{Strehl2002ClusterE} and Adjusted Rand Index (ARI)~\cite{Chacon2020MinimumAR}. 
To demonstrate the impact of the speaker diarization system, we will also report the Diarization Error Rate (DER)~\cite{Fiscus2006DER} with tolerance 0.25s and Jaccard Error Rate (JER)~\cite{Ryant2019TheSD}. The DER is generally composed of three parts: DER = Missed Speech (MS) + False Alarms (FA) + Speaker Error (SPKE). As the transcribed text and forced-alignment module have been used in the pipeline, we directly report the Concatenated Minimum-permutation Word Error Rate~\cite{Watanabe2020CHiME6CT}. Additionally, we use the metric Text Diarization Error Rate (TextDER)~\cite{Gong2023AligningSE}, to evaluate the amount of text assigned to wrong speakers. 

\section{Results and Discussion}

\subsection{Results of Speaker Diarization}

The results of speaker diarization in Table \ref{tab:constrained_sd} show that adding either semantic or visual constraints individually in E2CP + SC system can lead to improvements in cluster and diarization metrics. With semantic constraints, the JER metric decreases from 27.21\% to 25.98\%, and the DER metric also shows a reduction from 9.37\% to 9.12\%, compared to the baseline SC's DER. The integration of visual constraints improves cluster precision, as reflected by an improved NMI of 0.905.

The combination of semantic and visual constraints in E2CP + SC achieves the best performance across all evaluated metrics. It demonstrates a DER of 9.01\% and an NMI of 0.908, showcasing the synergistic effect of combining both types of constraints. TextDER and CpWER, which are from a textual perspective, also see significant improvement with the combined semantic and visual constraints. TextDER decreases from 3.25\% to 2.89\%, and CpWER from 17.04\% to 16.36\%. In our experiments, we keep the transcripts fixed, so the enhancement in CpWER is entirely attributable to the improvement of speaker diarization performance.

While relying solely on semantic constraints may appear to lag behind the use of visual constraints in clustering metrics, it has certain advantages in Speaker Metrics. Semantic constraints are often located near speaker change points, which can have a more substantial effect on the final diarization outcome, particularly for speaker identification within transcript text. On the other hand, visual information might suffer from asynchronicity during speaker change points, which can impede accurate speaker boundary determination. Therefore, the benefits of semantic information in adjudicating boundaries do not manifest in clustering metrics but confer a definitive advantage in speaker metrics.

\subsection{Constraint Construction and Analysis}


It is important to note that both visual and semantic decoding methods introduce some level of error, and the constraints constructed from these modalities may cover different sets of embedding pairs, so we need to examine the impact of these factors on the outcome. Table \ref{tab:constraint_constructions} presents the accuracy and coverage rates of constraints generated from both visual and semantic modalities.

It is evident that visual constraints outperform semantic constraints in terms of coverage. This difference can be attributed to the fact that the semantic tasks used in our model only evaluate relationships between embeddings within adjacent speaker turns, whereas visual constraints are assessed across embedding pairs with substantial temporal intervals. Furthermore, we have designed a method to effectively combine constraints from different modalities, and our findings indicate that semantic constraints provide a valuable supplement to visual constraints. In addition, we have conducted experiments in Appendix~\ref{appendix:constraints} to discuss the impact of constraints' quality and quantity on the results.

\subsection{Constraint Propagation Parameter}

As mentioned in Section \ref{jpcp}, $\lambda$ is a critical parameter during the constraint propagation process. By combining the analysis of Equations \ref{eq:propagation} and \ref{eq:change_affinity}, it can be found that when $\lambda$ tends towards 0, the final $\boldsymbol{\hat{\mathcal{Z}}}$ will be closer to $\boldsymbol{\mathcal{Z}}$, whereas when $\lambda$ approaches 1, the resulting $\boldsymbol{\hat{\mathcal{A}}}$ will be closer to $\boldsymbol{\mathcal{A}}$. 

Moreover, the parameter $\lambda$ also signifies the level of confidence that the model places in the constraints matrix. By adjusting the $\lambda$ value, the model can effectively handle different levels of error in the constraints, enabling the constrained propagation algorithm to adapt to models with varying performance. This adaptability is essential for effectively utilizing constraints in real-world scenarios. 

For additional details and insights on the impact of different $\lambda$ values on the algorithm's performance, please refer to the experiments illustrated in the Appendix~\ref{appendix:lambda}.

\subsection{Compare with Audio-visual Diarization}

\begin{table}
\centering
\caption{The results of audio-visual speaker diarization experiments on AVA-AVD datasets.}
\label{tab:ava}
\resizebox{\columnwidth}{!}{%
\begin{tabular}{cccccc}
\hline
                      & Modality        & Methods & VAD    & SPKE(\%)$\downarrow$  & DER(\%)$\downarrow$   \\ \hline
\multirow{2}{*}{AVR-Net}  & Audio          & VBx     & Oracle & 18.45 & 21.37 \\
                      & Audio + Visual & AVA-AVD & Oracle & 17.65 & 20.57 \\ \hline
\multirow{2}{*}{Ours} & Audio          & SC      & Oracle & 18.39 & 21.31 \\
                      & Audio + Visual & E2CP + SC    & Oracle & \textbf{17.40} & \textbf{20.32} \\ \hline
\end{tabular}%
}
\end{table}

In this section, we will demonstrate the effectiveness of our method on the audio-visual speaker diarization using the AVA-AVD~\cite{xu2022ava} dataset. Due to the lack of annotated  transcripts in the AVA-AVD dataset, in accordance with the original paper, only the SPKE and DER metrics are calculated. 

The method proposed in~\cite{xu2022ava} involves training an audio-visual relation network (AVR-Net) to simultaneously model audio and visual information. This approach requires a substantial amount of aligned audio-visual paired data for training. In contrast, our method utilizes a pretrained visual model to directly construct visual constraints that refine the affinity matrix from acoustic embeddings. Table~\ref{tab:ava} presents a comparison conducted on AVA-AVD, revealing the competitive performance of our method when utilizing only visual constraints. This demonstrates strong generalizability of our approach.

\subsection{Compare with Audio-textual Diarization}

In this section, we will compare our method with existing audio-text speaker diarization approaches, specifically evaluating our performance against the system presented in \cite{Luyao2023ACL} using the AIShell-4 and Alimeeting datasets. Since both datasets are in Mandarin, we utilize the open-sourced speaker embedding model in the same manner as their work for a fair comparison. To mitigate the influence of the ASR system, our experiments will focus on the results decoded from the ground-truth annotated text, as described in their work. Consistent with their findings, we will report the metrics CpWER and TextDER.


The method presented in \cite{Luyao2023ACL} integrates acoustic and semantic information to address boundary issues in acoustic-only systems. In contrast, our approach uses semantic information to create pairwise constraints that directly impact spectral clustering. Table \ref{tab:audio-text} presents the experimental results of our audio-text speaker diarization method, showing significant performance improvements on both the AIShell-4 and AliMeeting datasets. Specifically, the CpWER on the AIShell-4 dataset decreased from 15.23\% to 14.95\%, and on the AliMeeting dataset, it reduced from 36.15\% to 31.11\%. These results indicate that our method effectively utilizes semantic information compared to the approach in \cite{Luyao2023ACL}.

\begin{table}[]
\centering
\caption{The results of audio-text speaker diarization experiments on AIShell-4 and Alimeeting Datasets.}
\label{tab:audio-text}
\resizebox{\columnwidth}{!}{%
\begin{tabular}{ccccc}
\hline
Dataset                     & Modality         & Methods           & CpWER(\%)$\downarrow$ & TextDER(\%)$\downarrow$ \\ \hline
\multirow{3}{*}{AIShell-4}  & Audio            & SC                & 17.31     & 5.97        \\
                            & Audio + Semantic & Fusion & 15.23     & 6.28        \\
                            & Audio + Semantic & Ours              & \textbf{14.95}     & \textbf{4.98}       \\ \hline
\multirow{3}{*}{Alimeeting} & Audio            & SC                & 41.67     & 18.89       \\
                            & Audio + Semantic & Fusion & 36.15     & 14.50       \\
                            & Audio + Semantic & Ours              & \textbf{31.11 }    & \textbf{10.76}      \\ \hline
\end{tabular}%
}
\end{table}

\section{Conclusions}
In this study, we propose a novel multimodal approach that jointly leverages audio, visual, and semantic information for enhanced speaker diarization. Additional visual and textual processing modules are incorporated to generate complementary visual and semantic constraints. A joint pairwise constraint propagation method is employed to integrate multimodal information into the speaker clustering process. Experimental results confirm the significant improvement in diarization performance.

We posit that our study represents a significant advancement in the field of multimodal speaker diarization. By incorporating an augmented array of modal information, we provide a framework that not only enriches the current understanding of speaker diarization processes but also catalyzes further innovation and exploration within this domain. Our work sets the stage for the development of more sophisticated systems that can accurately parse and attribute speaker identity, thereby broadening the horizon for future research endeavors.

\appendix

\bibliography{aaai25}

\section*{Appendix}

\begin{figure*}
    \centering
    \includegraphics[width=0.96\textwidth]{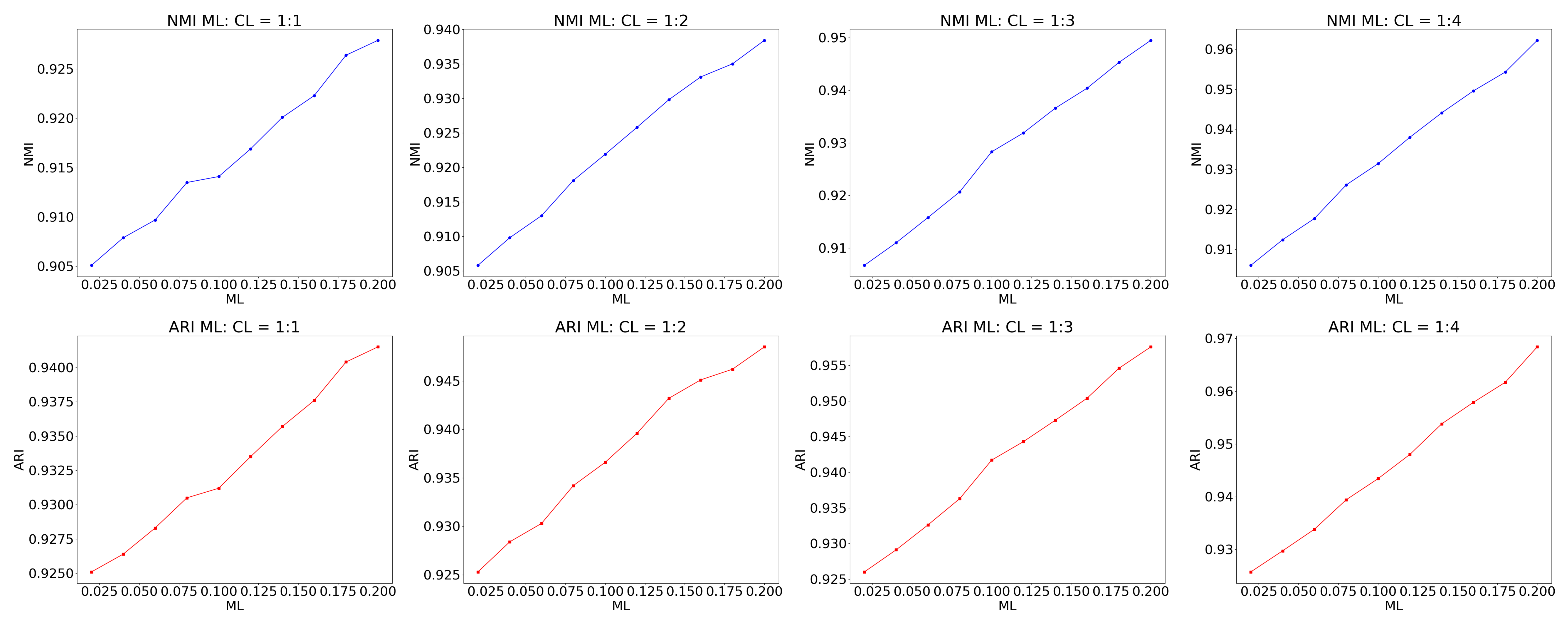}
     \caption{Results of constrained speaker cluster performance across various levels of constraints coverage, showcasing scenarios with imbalanced proportions of must-link and cannot-link constraints.}
    \label{fig:coverage_cluster}
\end{figure*}

\section{Constraints Sensitivity Analysis}
\label{appendix:constraints}

In this section, we discuss the effects of constraint quality and quantity on experimental results. To facilitate variable control within our experiments, we employed a batch of simulated constraints. Specifically, the actual speaker labels for each extracted speaker embedding were determined using ground truth speaker timestamps, and the corresponding constraints were constructed by comparing these ground truth labels.

\subsection{The Impact of Constraint Quantity}

For a sequence of speaker embeddings $E = \{e_1, e_2, ..., e_N | e_i \in \mathbb{R}^D\}$, it should satisfy $|\mathcal{M}| + |\mathcal{C}| \le N \times (N - 1)$. Given that the number of speakers in a meeting scenario should be greater than or equal to 2, generally, $|\mathcal{C}| > |\mathcal{M}|$. To assess the impact of the quantity of constraints and the imbalance between must-link and cannot-link, we employ the following strategy for randomization: First, we randomly determine the must-link coverage coefficient $p_{ml}$ and the cannot-link coverage coefficient $p_{cl}$, where $p_{ml} \in \{2\%, 4\%, 6\%, ..., 20\%\}$ , and $p_{cl} = k_{imbalance} \times p_{ml}$ with $k_{imbalance} \in \{1, 2, 3, 4\}$. Ultimately, we select a proportion $p_{ml}$ of must-links from all possible $\mathcal{M}$ and a proportion $p_{cl}$ of cannot-links from all possible $\mathcal{C}$ to form a set of constraints.

Our experimental findings are presented in Figure \ref{fig:coverage_cluster}. We observed that regardless of the ratio of constraint types, whether must-link or cannot-link (ranging from 1:1 to 1:4), there is a definitive enhancement in clustering performance as the constraints encompass an increasing number of embedding pairs.

\subsection{The Impact of constraint Quality}

\begin{figure}[ht]
    \centering
    \includegraphics[width=0.47\textwidth]{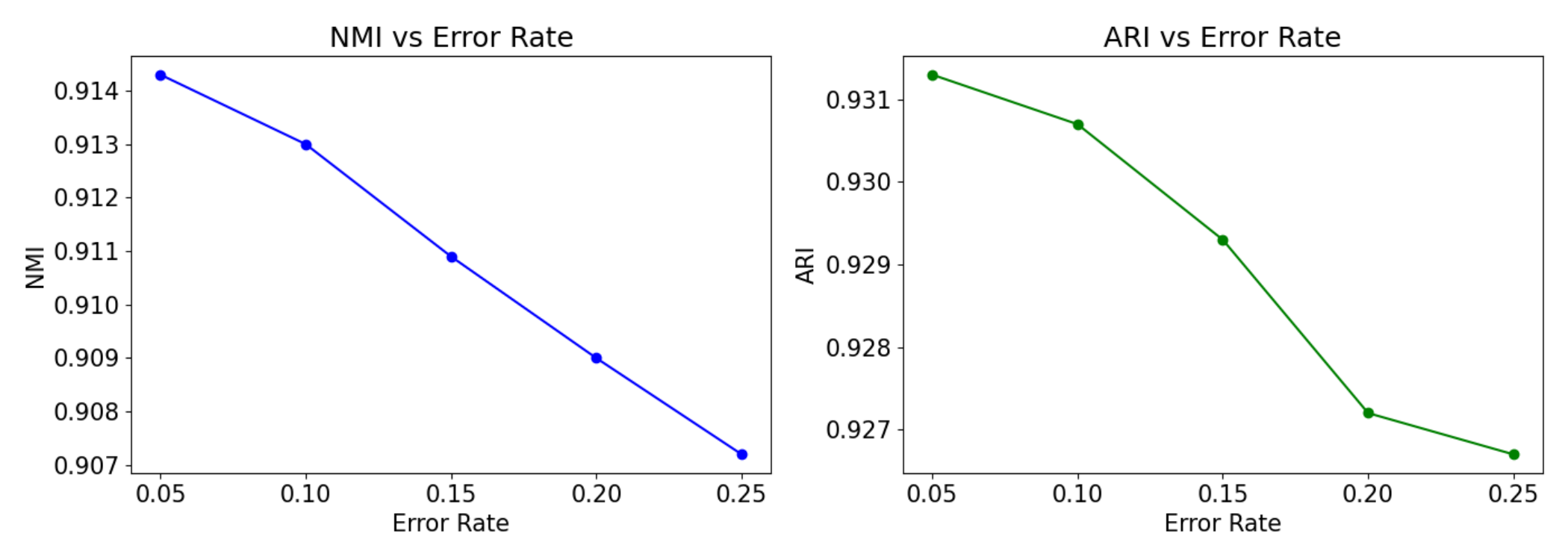}
     \caption{Simulated constraints with errors and the effect for constrained clustering}
    \label{fig:error_cluster}
\end{figure}

Nearly all pairwise constrained clustering methods assume that the input constraints are entirely accurate; however, in practice, the constraints we obtain often contain many errors. This is especially common in multi-party meeting or interview scenarios, such as when there is audio-visual asynchrony or errors from transcipt text decoded by ASR due to complex acoustic environments. In order to investigate the impact of incorrect constraints on our method, we have established the following randomization strategy: First, we randomly generate a completely correct set of constraints, including must-links and cannot-links. We then randomly alter the status of a proportion $p_{err}$ of these constraints—turning must-links into cannot-links and vice-versa—thereby introducing a certain level of constraint errors while keeping the total number of constraints constant. In our experiments, $p_{err} \in \{5\%, 10\%, 15\%, 20\%, 25\%\}$.

The related results are reported in Figure \ref{fig:error_cluster}, where it can be observed that erroneous constraints significantly degrade the clustering outcomes. This implies that utilizing our multimodal framework will benefit from the enhancement of multimodal model capabilities.

\section{Constrained Cluster Parameters Analysis}
\label{appendix:lambda}
\begin{figure}
    \centering
    \includegraphics[width=0.47\textwidth]{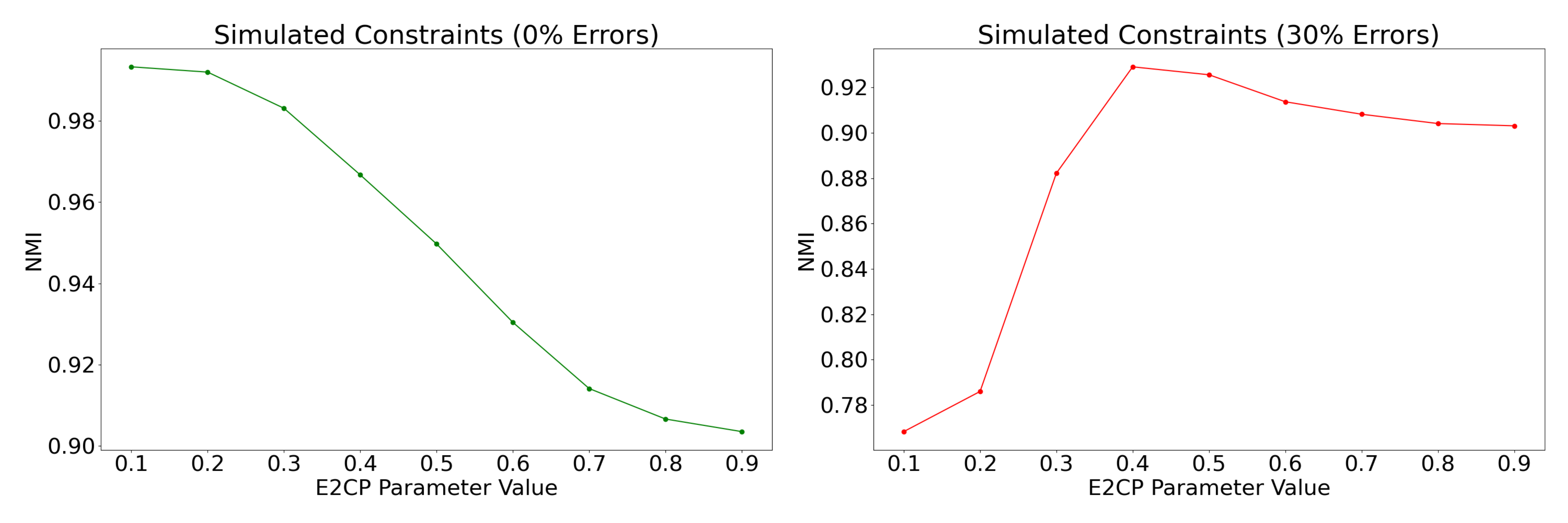}
     \caption{Analysis of constrained clustering outcomes with varying $\lambda$ values. It is observed that when constructed constraints contain errors, the peak of the optimal $\lambda$ shifts towards 1.0.}
    \label{fig:e2cp}
\end{figure}

We conducted simulations of constraints to compare the optimal $\lambda$ values when introducing errors in the constraints. The Figure \ref{fig:e2cp} illustrate that the optimal E2CP parameter value $\lambda$ for maximizing NMI depends on the error rate within the constraints. With $0\%$ errors, the best performance is achieved at the lowest $\lambda = 0.1$, indicating that with highly accurate constraints, the algorithm benefits from a strong adherence to constraint guidance. However, for constraints with a $30\%$ error rate, the peak NMI occurs at a higher $\lambda = 0.4$, suggesting that with less reliable constraints, the algorithm requires a more moderate constraint influence to balance error tolerance and performance. These results highlight the importance of adjusting $\lambda$ in accordance with the fidelity of constraints to achieve optimal speaker diarization.

\end{document}